# Euler angles based loss function for camera relocalization with Deep learning

Qiang Fang , Tianjiang Hu

Abstract--Deep learning has been applied to camera relocalization, in particular, PoseNet and its extended work are the convolutional neural networks which regress the camera pose from a single image. However there are many problems, one of them is expensive parameter selection. In this paper, we directly explore the three Euler angles as the orientation representation in the camera pose regressor. There is no need to select the parameter, which is not tolerant in the previous works. Experimental results on the 7 Scenes datasets and the King's College dataset demonstrate that it has competitive performances.

## 1. Introduction

Designing a system for pose estimation based on a single camera is a challenging system, which is important in many fields, such as robotic navigation and augmented reality, etc.

In the past few decades, many methods are proposed[1-6] for solving the problem, but those methods are unable to tackle all the scene, especially the extreme conditions. Recently, the approaches based on convolutional neural networks (CNNs) have been developed to perform camera relocalization due to their adaptive capabilities to illumination change and motion blur.

The CNN-based methods have achieved good results on both the 7

scenes indoor dataset and the large scale outdoor datasets, however, there are still open problems. The proposed deep learning models[7-9] regress the translation and the orientation(quaternion-based or the trigonometric function of the Euler angle) simultaneously, which have the inherent problem for the nondimensionalize of the orientation representation. It requires a parameter to balance the translation and orientation, however the parameter differs greatly for different datasets or different CNN models, which is not tolerant to the selection of the parameter, and there might not be a common skill to select the parameter.

Our motivation in this paper is to tackle the problem stated above, following the intuition that the dimension of Euler angles is degree and that of the translation is meter, we directly adopt the three dimensional Euler angles as the orientation representation in the camera pose regressor. We can change the confidence by changing the weighting factor, for instance, if the weighting factor is setted as 1, that means regressing the translation and orientation into 1 meter and 1 degree respectively. The method is easy, and its suitability is wide, what is more, experimental results showed that it has competitive performance.

## 2. Related work

There are generally three approaches for vision based camera relocalization: feature-based, pixel-based and learning-based approaches.

1) **Feature-based approaches**

Feature based approaches firstly detect interest points in the image, extract their local features,such as SIFT [10] and ORB [11], match(or track) them against a database of features, and then employ multi-view geometry theory [12] to regress camera pose using the register points[13]. However, the feature based algorithms suffer from drifts over time. To mitigate this problem, visual simultaneous localisation and mapping (SLAM) is adopted for drift correction along with pose estimation, such as PTAM[14] and ORB-SLAM[15], however, these methods are computationally expensive.

2) **Pixel-based approaches**

Feature based approaches only use features without other useful informations contained in the image. Pixel-based approaches(or direct approaches), in contrast, utilize all the pixels in consecutive images for pose estimation under the assumption of photometric consistency. Recently, DTAM and semi-direct approaches which achieve better performance are developed[16-19], especially in texture-less environments, however, these methods are still suffering computationally expensive.

3) **Learning-based approaches**

Deep learning has achieved exciting results on the fields of image detection, segmentation and classification, however convolutional neural networks are only just beginning to be used for camera pose regression[20]. SCoRe Forests[21] use random forests to regress scene coordinate labels to relocalization . However, this algorithm is limited to indoor scenes due to the need of RGB-D images. A CNN-based relocalization framework named PoseNet is introduced[7], it can globally relocalise without a good initial pose estimate, and produces a continuous metric pose. Many extended work have been proposed, such as interpret relocalisation uncertainty with Bayesian Neural Networks[8] and the Euler6 method[9], additionally, [22] demonstrate PoseNet's efficacy on featureless indoor environments, where the SIFT based techniques fail. However these methods require expensive parameter selection, in order to avoid the problem,[23] explore a loss functions for learning camera pose which are based on geometry and scene reprojection error, however a good initial weight is needed during the training process.

## 3. Model for deep regression of camera pose

### 3.1 Loss function

Our task is to estimate camera pose directly from a monocular image, the outputs are two vectors respectively representing orientation and translation. In this section, we present the novel orientation representation Euler angles $\Phi = [\psi, \theta, \phi]$, which are three angles(yaw,pitch and roll) to describe the orientation of a rigid body. They represent a sequence of rotations about the axes of a coordinate system. Euler angles face a similar problem with quaternion: the periodicity leads to quite different angle values for similar images. To address this, we constrain all angles to an interval $(-\pi, \pi]$ in the euclidean layer.

To regress pose, we adopt the GoogLeNet like the PoseNet but we modify the the objective loss function as follows:

$$Loss(I) = w_1 \left\| \hat{X} - X \right\|_2 + w_2 \left\| \hat{\Phi} - \Phi \right\|_2 \qquad (1)$$

Where parameters $w_1, w_2$ are the weights of the translation and orientation respectively. $X = [x, y, z]$ is the translation vector and $\hat{X}$ is the corresponding labeled data, $\hat{\Phi}$ is the Euler angles' labeled data. Our purpose is to train the loss function (1) to reach the minimum(or convergence).

## 3.2 Criterion of angle error

In the PoseNet method, the orientation is represented using quaternion $q = [q_w, q_x, q_y, q_z]$, then the quaternion error $\delta q$ can be written as

$$\delta q = q \circ \hat{q} = [\delta q_w, \delta q_x, \delta q_y, \delta q_z] \qquad (2)$$

where $\hat{q} = [\hat{q}_w, \hat{q}_x, \hat{q}_y, \hat{q}_z]$ is the labeled quaternion and '∘' is the multiplication of quaternions, and the angle error $\Delta$ is defined as

$$\Delta = 2 \arccos(\delta q_w) \qquad (3)$$

However in this paper, the orientation is represented by Euler angles $\Phi = [\psi, \theta, \phi]$, so for comparison we first change the Euler angles into quaternion, and then obtain the angle error using (2) and (3).

## 3.3 Median or Mean

In the PoseNet method, all the results are evaluated with median, however median might not be a good evaluation. Figure 1 shows an example, there are 10 data, and the median is 1.1 ,while it is obviously not representative, on the contrary, mean(here it is 1.4778 ) is a better choice, so in this paper , we also added the mean results for comparison.

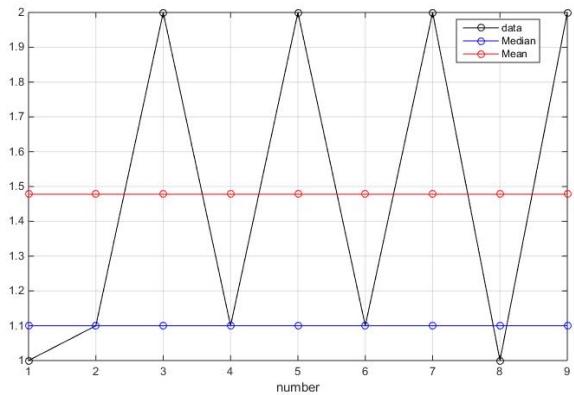

Fig.1 Median vs Mean

## 4. Experiment

The proposed methods were evaluated on an indoor relocalization dataset 7 Scenes[24] and an outdoor dataset(King's College)[8]. All experiments were based on Caffe [25].

The models are trained end-to-end using stochastic gradient descent. the parameters $w_1, w_2$ were set to $w_1 = 1, w_2 = 1$. We train each model until the training loss converges using the default parameters and a learning rate of $10^{-3}$. We use a batch size of 64 on a NVIDIA GTX 1080 GPU, training takes approximately 30k-200k iterations, or about 7 hours or 2 days.

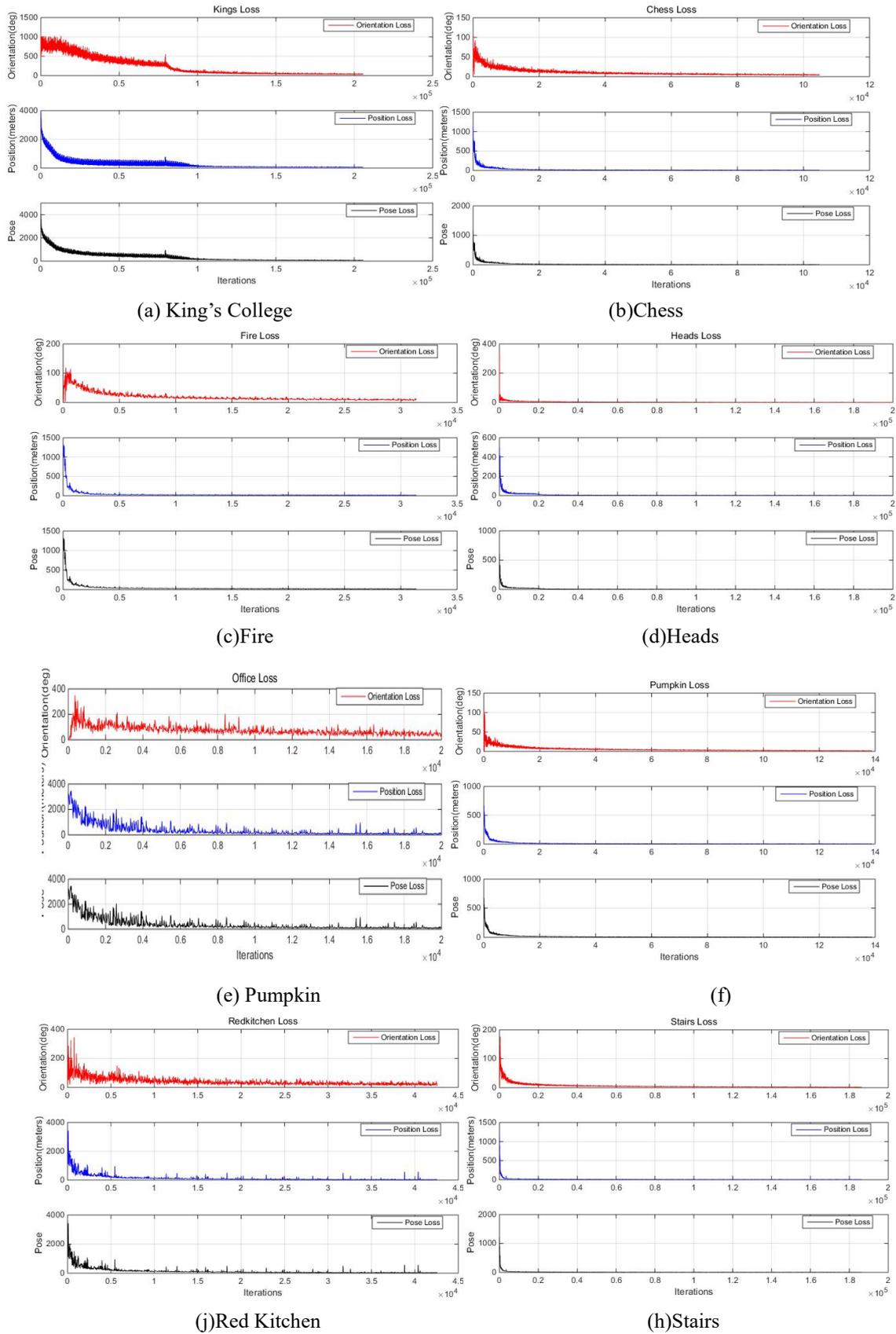

(a) King's College  (b)Chess
(c)Fire  (d)Heads
(e) Pumpkin  (f)
(j)Red Kitchen  (h)Stairs

Fig.2 Training loss among different scenes.

Figure 2 shows the training losses of the indoor and outdoor datasets.

During the whole process of training, there is no need to adjust the parameters as the PoseNet does, we just need to judge whether the training is convergent. As can be observed from the plots, all the curves converge after different numbers of iterations, when we stopped the training.

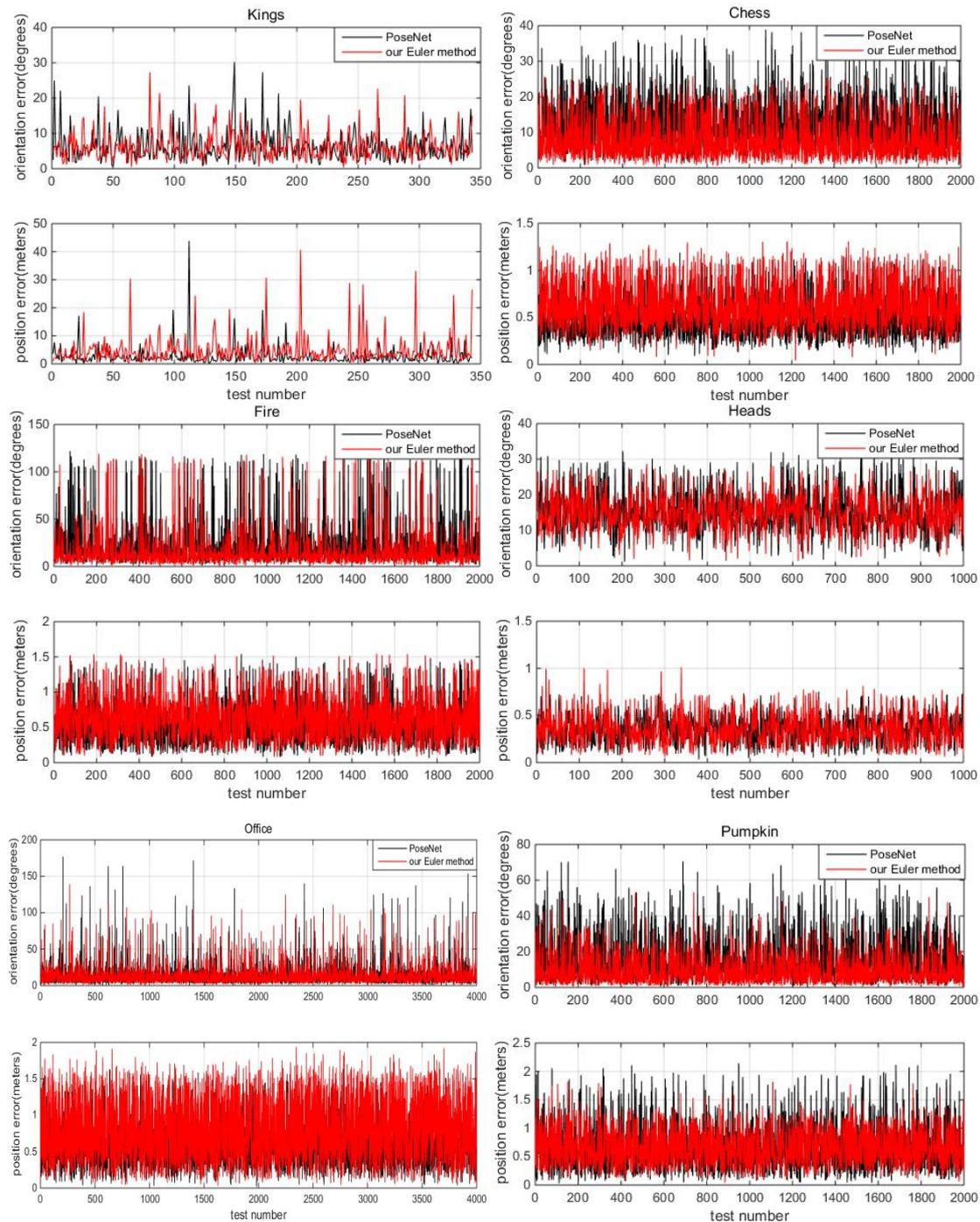

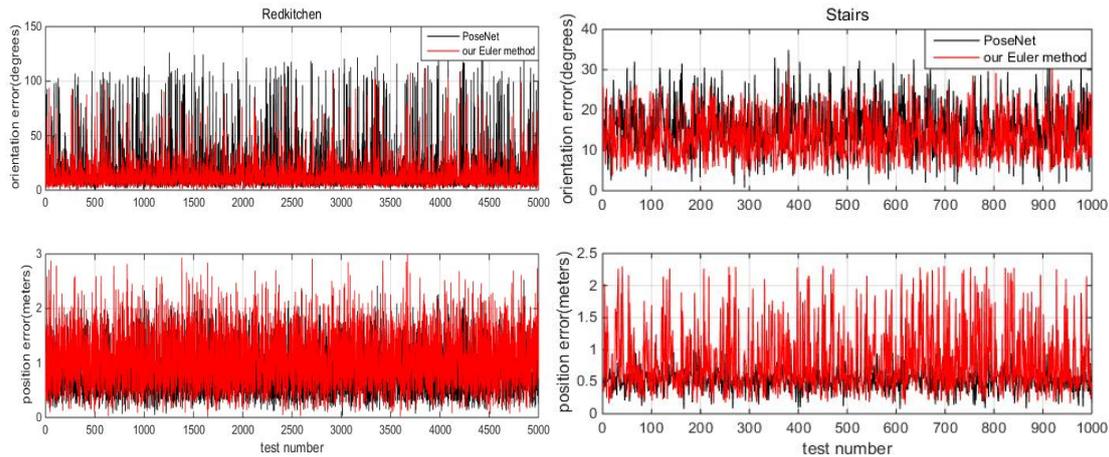

Fig.3 Orientation and position errors among of the test frames

Figure 3 shows the orientation and position errors of the indoor and outdoor datasets during the whole process of testing. It can qualitatively conclude that the orientation errors of this paper outperform the PoseNet, and the translation errors of this paper are almost the same levels with the PoseNet, except in the outside scene, which is a little higher.

Table 1 Median and mean errors of different scene

| Scene | Frames | | Median | | Mean | |
|---|---|---|---|---|---|---|
| | Train | Test | PoseNet | Our method | PoseNet | Our method |
| King's College | 1220 | 343 | 1.92m, 5.40° | 3.5714m, **5.2756°** | 2.6961m, 6.4132° | 4.9158m, **6.2936°** |
| Chess | 4000 | 2000 | 0.32m, 8.12° | 0.5623m, **5.8011°** | 0.4709m, 12.3897° | 0.6208m, **7.7281°** |
| Fire | 2000 | 2000 | 0.47m, 14.4° | 0.6362m, **9.7375°** | 0.5413m, 23.4793° | 0.6431m, **15.4148°** |
| Heads | 1000 | 1000 | 0.29m, 12.0° | 0.3358m, 14.6991° | 0.3462m, 15.0283° | 0.3562m, **14.9700°** |
| Office | 6000 | 4000 | 0.48m, 3.84° | 0.7441m, 9.9039° | 0.5770m, 13.0190° | 0.8054m, 13.7793° |
| Pumpkin | 4000 | 2000 | 0.47m, 8.42° | 0.6343m, **6.3086°** | 0.6491m, 15.2436° | 0.6746m, **9.6233°** |
| RedKitchen | 7000 | 5000 | 0.59m, 8.64° | 0.9794m, 9.7126° | 0.7656m, 17.2293° | 1.0523m, **13.0946°** |
| Stairs | 2000 | 1000 | 0.47m, 13.8° | 0.6153m, **11.5980°** | 0.5213m, 15.8150° | 0.8156m, **13.1620°** |

Table 1 shows the median and mean performances for the 7 Scenes datasets and the King's College dataset using our Euler angles based loss function, in addition, the PoseNet is also shown for comparison. Experimental results showed that the translation is a little higher both in the median error and mean error, especially for the outdoor scene, but our method outperformed PoseNet in some ways. Most of the orientation results are lower than the PoseNet, especially the mean error of the Chess dataset has a 37.6% reduction over PoseNet. This demonstrates that our novel loss function is competitive and is an alternative for camera pose regression compared to the PoseNet.

## 5. Conclusion

We have investigated a loss function for learning to regress pose with Euler angles. Although the accuracy of the translation results is not as high as the PoseNet, the accuracy is not much lower and all the mean errors of the orientation is better than PoseNet. More importantly, the present algorithm does not require any hyper-parameter tuning, which make it more common for different datasets(or different networks) than PoseNet during the training process.

**Acknowledgement**

This work was supported by the National Science Foundation for Young Scientists of China (Grant No:61703418) .